\documentclass[letterpaper, 10 pt, conference]{ieeeconf}  
\usepackage{graphicx}
\usepackage{amsmath}
\usepackage{mathrsfs}
\usepackage[style=ieee]{biblatex}

\usepackage{xcolor}
\usepackage{balance}
\DeclareSourcemap{
  \maps{
    \map{
      \pertype{article}
      \step[fieldset=url, null]
      \step[fieldset=doi, null]
      \step[fieldset=issn, null]
      \step[fieldset=isbn, null]
      \step[fieldset=note, null]
      \step[fieldset=editor, null]
      \step[fieldset=urldate, null]
      \step[fieldset=file, null]
    }
  }
}
\DeclareSourcemap{
  \maps{
    \map{
      \pertype{inproceedings}
      \step[fieldset=url, null]
      \step[fieldset=doi, null]
      \step[fieldset=issn, null]
      \step[fieldset=isbn, null]
      \step[fieldset=note, null]
      \step[fieldset=editor, null]
      \step[fieldset=urldate, null]
      \step[fieldset=file, null]
    }
  }
}
\DeclareSourcemap{
  \maps{
    \map{
      \pertype{incollection}
      \step[fieldset=url, null]
      \step[fieldset=doi, null]
      \step[fieldset=issn, null]
      \step[fieldset=isbn, null]
      \step[fieldset=note, null]
      \step[fieldset=editor, null]
      \step[fieldset=urldate, null]
      \step[fieldset=file, null]
    }
  }
}
\DeclareSourcemap{
  \maps{
    \map{
      \pertype{misc}
      \step[fieldset=url, null]
      \step[fieldset=doi, null]
      \step[fieldset=issn, null]
      \step[fieldset=isbn, null]
      \step[fieldset=note, null]
      \step[fieldset=editor, null]
      \step[fieldset=urldate, null]
      \step[fieldset=file, null]
    }
  }
}

\IEEEoverridecommandlockouts                              

\title{\LARGE \bf
Vision-Guided Loco-Manipulation with a Snake Robot
}

\author{Adarsh Salagame$^{1\text{\textdagger}}$, Sasank Potluri$^{1\text{\textdagger}}$, Keshav Bharadwaj Vaidyanathan$^{1\text{\textdagger}}$, \\ Kruthika Gangaraju$^{1}$, Eric Sihite$^{2}$, Milad Ramezani$^{3*}$, Alireza Ramezani$^{1}$
\thanks{$^{1}$This author is with the Department of Electrical and Computer Engineering, Northeastern University, Boston MA
        {\tt\small salagame.a, potluri.sas, bharadwajvaidyanat.k, gangaraju.k, a.ramezani@northeastern.edu*}}%
\thanks{$^{2}$ This author is with California Institute of Technology, Pasadena CA
		{\tt\small esihite@caltech.edu}}%
\thanks{$^{3}$Author is with CSIRO Robotics, DATA61, CSIRO, Brisbane, Australia. Email: 
        {\tt\small milad.ramezani@data61.csiro.au}}%
\thanks{\textdagger These authors have equal contribution to this work.}
\thanks{$*$Indicates the corresponding author.}
}

\begin{document}

\maketitle
\thispagestyle{empty}
\pagestyle{empty}

\begin{abstract}

This paper presents the development and integration of a vision-guided loco-manipulation pipeline for Northeastern University's snake robot, COBRA. The system leverages a YOLOv8-based object detection model and depth data from an onboard stereo camera to estimate the 6-DOF pose of target objects in real time. We introduce a framework for autonomous detection and control, enabling closed-loop loco-manipulation for transporting objects to specified goal locations. Additionally, we demonstrate open-loop experiments in which COBRA successfully performs real-time object detection and loco-manipulation tasks.

\end{abstract}

\section{Introduction}

Limbless animals like snakes can traverse complex terrain. These biological systems have inspired roboticists for decades \cite{marvi_sidewinding_2014, liu_locomotion_2024, ramesh_sensnake_2022, wang_cpg-inspired_2017, bing_towards_2017}. Prevalent research in snake robotics has been primarily concerned with locomotion. The over-actuated morphology of snake robots makes it powerful for loco-manipulation tasks as well which is explored to some extend \cite{reyes_planar_2014, sanfilipp_combining_2022}.

Loco-manipulation involves kinematics and dynamics modeling, path planning, object detection, localization, etc., and has been extensively studied in the context of manipulators attached to traditional mobile robots such as quadrupeds \cite{bellicoso_alma_2019, chen_design_2022}, wheeled robots \cite{carius_deployment_2018, bischoff_kuka_2011,sihite_multi-modal_2023}, aerial robots \cite{jimenez-cano_aerial_2015, kim_globally_2023}.

Research on topics such as searching-exploration strategies, looking for target - mapping and exploration \cite{lluvia_active_2021}, target detection by assuming target is within field of view \cite{ren_faster_2017, liu_deep_2020}, pose identification and identifying 6-dof pose (position and orientation) \cite{wohlhart_learning_2015, grabner_3d_2018}, affordance detection and identify ways of manipulating object \cite{kaiser_towards_2016, kokic_affordance_2017, pohl_affordance-based_2020}, grasping target \cite{du_vision-based_2021, reyes_planar_2014}, and navigation and locomotion \cite{murooka_humanoid_2021}
have formed a rich array of research opportunities under loco-manipulation field.

Snake robots possess manifold actuated joints that can be assigned for locomotion and manipulation tasks. As a result, the above topics can take excitingly unexplored forms. In addition, from robot capability stand point, they can operate inside confined spaces, traverse rugged terrain and reach manipulation targets that cannot be reached by these robots, and still retain the dexterity needed to manipulate the position and orientation of detected objects. This capability makes them increasingly useful for hostile environments such as operations on Moon and Mars lava tubes, caves, craters, etc., \cite{marvi_sidewinding_2014} 
search and rescue in the aftermath of natural disasters that yield highly unstructured terrain and task space \cite{whitman_snake_2018}. 

\begin{figure}
    \centering
    \includegraphics[width=0.9\linewidth]{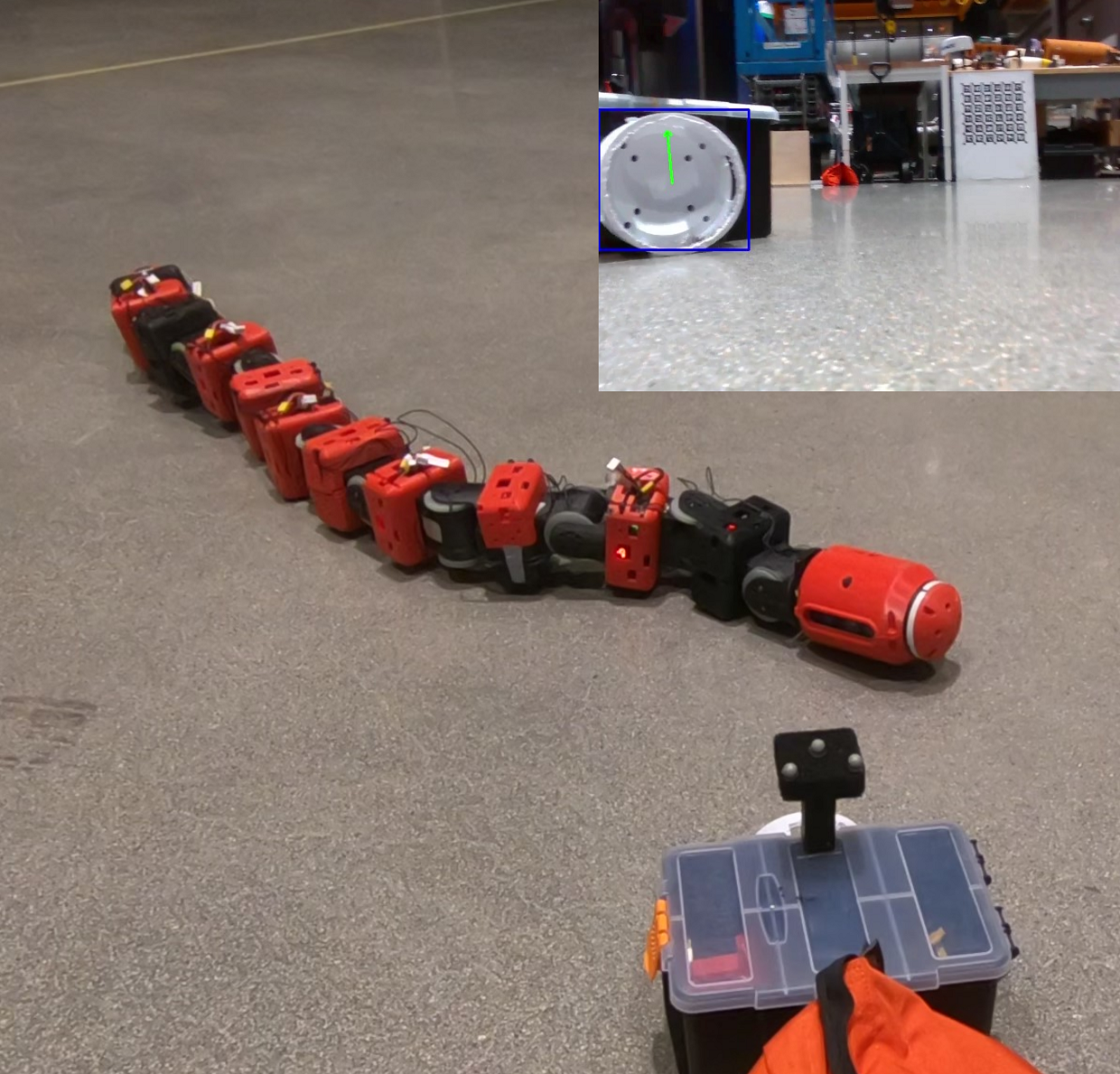}
    \caption{Shows COBRA \cite{salagame_validation_2024,salagame_loco-manipulation_2024,salagame_reinforcement_2024,salagame_how_2024} sidewinding towards target for loco-manipulation (Inset: View from onboard camera with detection of target object)}
    \label{fig:cover}
    \vspace{-5mm}
\end{figure}

In our previous works \cite{salagame_non-impulsive_2024, salagame_loco-manipulation_2024}, we presented our preliminary result on a framework for contact implicit path planning for loco-manipulation and demonstrated hardware and simulation results for transporting objects. These research involved utilizing controlled snake body motions to push objects on flat ground. 

In this work, however, we shift our focus on exploiting morpho-funcationality of our snake robot for successful target detection and pose identification from distance. Joint redundancy can be levereged to postion detection sensors in different poses and therefore enhance detection. For instance, in our COBRA robot the gaits can be designed so that no matter how the camera is fixated, it faces to object as the robot sidewinds. 

We seek implementation of the following pipeline in this paper:

\begin{enumerate}
    \item Considering the target (shown in Fig.~\ref{fig:cover}) exists in the task space of COBRA, the robot detects the docking module using its onboard camera while performing sidewinding locomotion,
    \item then, COBRA, fully relied on it onboard electronics, generates and uses depth information from stereo-camera and robot localization to estimate the 6-DoF pose of the docking module in real-time,
    \item After, COBRA uses the estimated docking module pose as a reference to perform closed-loop docking with the target box, effectively picking up the object by autonomously applying its docking mechanism.
    \item Last, COBRA transport the box to a designated location through trajectory planning and path tracking fully relied on it onboard computer.
\end{enumerate}

\begin{figure}
    \centering
    \includegraphics[width=0.9\linewidth]{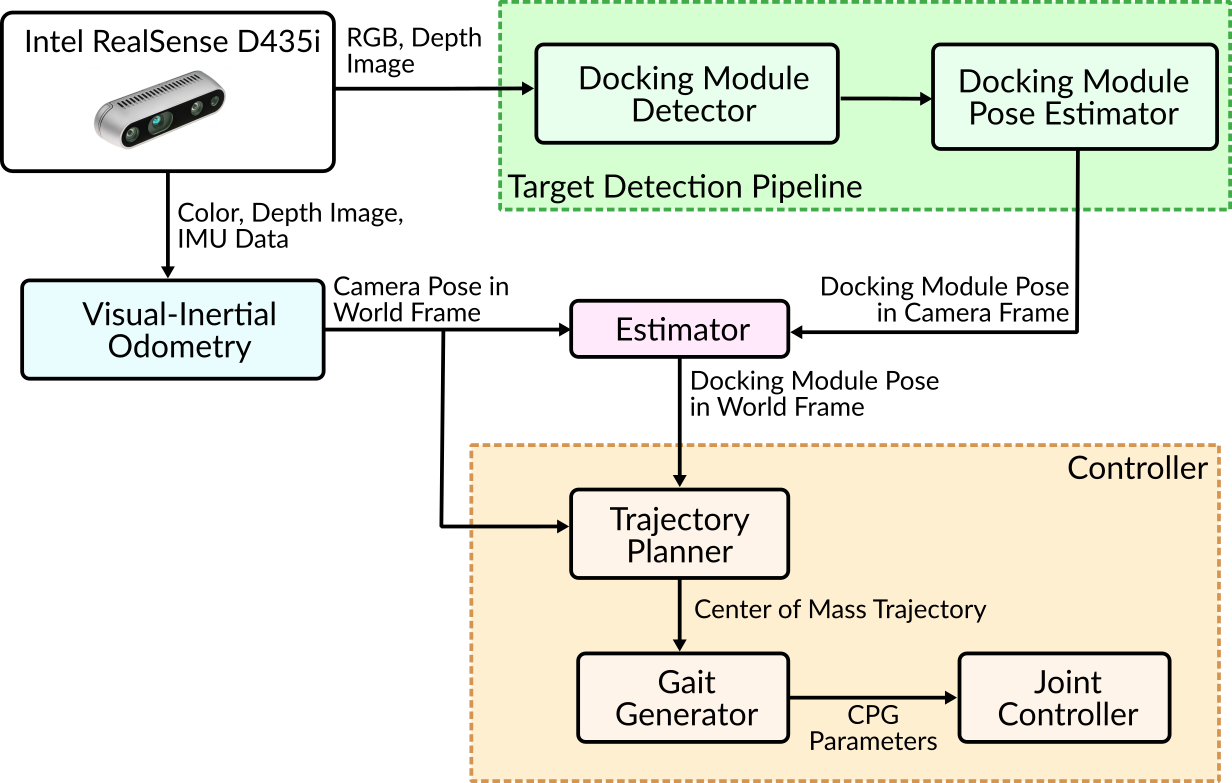}
    \caption{Full proposed system pipeline}
    \label{fig:system-overview}
    \vspace{-5mm}
\end{figure}

Figure \ref{fig:system-overview} illustrates the full proposed pipeline for closed-loop loco-manipulation on CORBA. The main contribution of this work lies in system integaration and demonstration of full pipeline on autonomous object detection and manipulation. Hence, at hardware design level we identified sensors and computers that can successfully be hosted by COBRA and substantiate the tasks considered. At software level, we used state-of-the-art segmentation, bounding box detection, and object approach direction estimation to serve our autonomy objectives. 

In this work, we implement the detection and pose estimation components, and propose the control framework for closed-loop docking. We also present initial results showcasing the onboard perception pipeline and demonstrate open-loop docking and loco-manipulation as a proof of concept.

This work is organized as follows. Section. \ref{sec:overview} gives a brief overview about the COBRA platform. Section \ref{sec:detection} discusses the approaches taken towards detecting the box and gives performance evaluations of the chosen approach. Section \ref{sec:dock-pose-estim} presents a framework for estimating the 6-degree-of-freedom (6-DoF) pose of the box once detected, and Section \ref{sec:control} provides a high level controls framework close the loop to perform autonomous loco-manipulation. Finally, Section \ref{sec:res} presents results for the detection and pose estimation pipeline and an open loop hardware test demonstrating a full loco-manipulation pipeline.


\section{Overview of COBRA Platform}
\label{sec:overview}

This section briefly covers three hardware aspects of COBRA: body design, docking mechanism, perception unit.

\subsection{Body Design}
COBRA \cite{jiang_snake_2024,jiang_hierarchical_2024,salagame_how_2024} is comprised of eleven actuated joints and twelve links, starting with a head module housing an Nvidia Jetson Orin NX processor, and an Intel RealSense D435i stereo-camera with an inertial measurement unit (IMU). The rear section houses an interchangeable payload module, accommodating additional electronics tailored to specific tasks undertaken by COBRA. The remaining components consist of identical modules, each housing a joint actuator and a battery.


\subsection{Docking Mechanism}
\begin{figure*}
    \centering
    \includegraphics[width=0.6\linewidth]{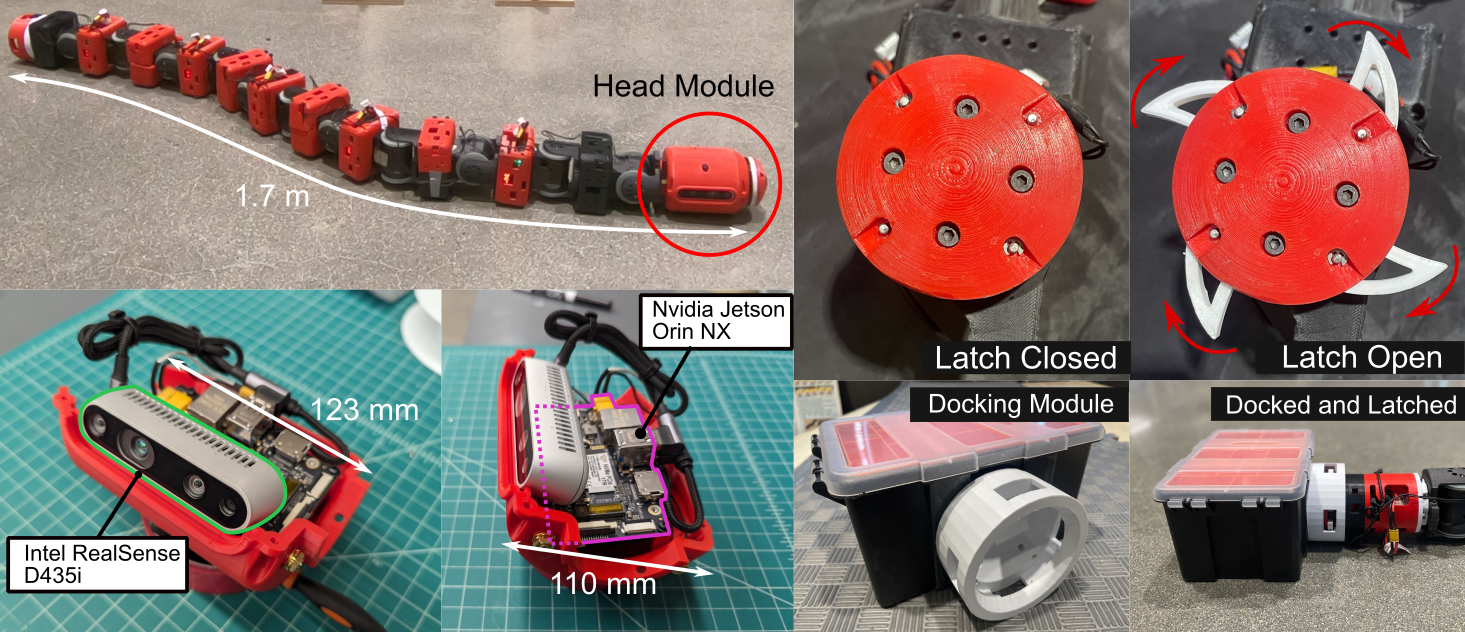}
    \caption{(Above Right) Closeup view of the head with actuated fins. (Below Right) Docking module attached to object for loco-manipulation. (Below Left) Shows NVidia Jetson Orin NX and Intel RealSense D435i mounted in the head module}
    \label{fig:head}
\end{figure*}
A latching mechanism (see Fig.~\ref{fig:head}) is integrated into the head module that functions as a multi-purpose gripper used for COBRA's tumbling locomotion \cite{salagame_how_2024} as well as for object manipulation tasks \cite{salagame_loco-manipulation_2024}. The latch is driven by a Dynamixel XC330 actuator situated within the head module, driving a central gear that opens and closes four latching fins. These are used to grip onto a complementary module called the "Docking Module" attached to the target object. 

\subsection{Perception Unit}
COBRA's head, in addition to the docking mechanism explained above, hosts perception unit. This unit is composed of an Nvidia Jetson Orin NX processor with 8GB of RAM that is used for high level autonomy tasks such as mapping and localization as well as for the core low level controls of the robot, operating a control loop of 500 Hz. It also contains an Intel RealSense D435i stereo-camera with an Inertial Measurement Unit (IMU) that provides time synchronized and aligned RGB and depth images along with time synchronized IMU measurements. This is used for state estimation for closed-loop locomotion as well as for high level planning tasks such as loco-manipulation.

\begin{figure}
    \centering
    \includegraphics[width=1\linewidth]{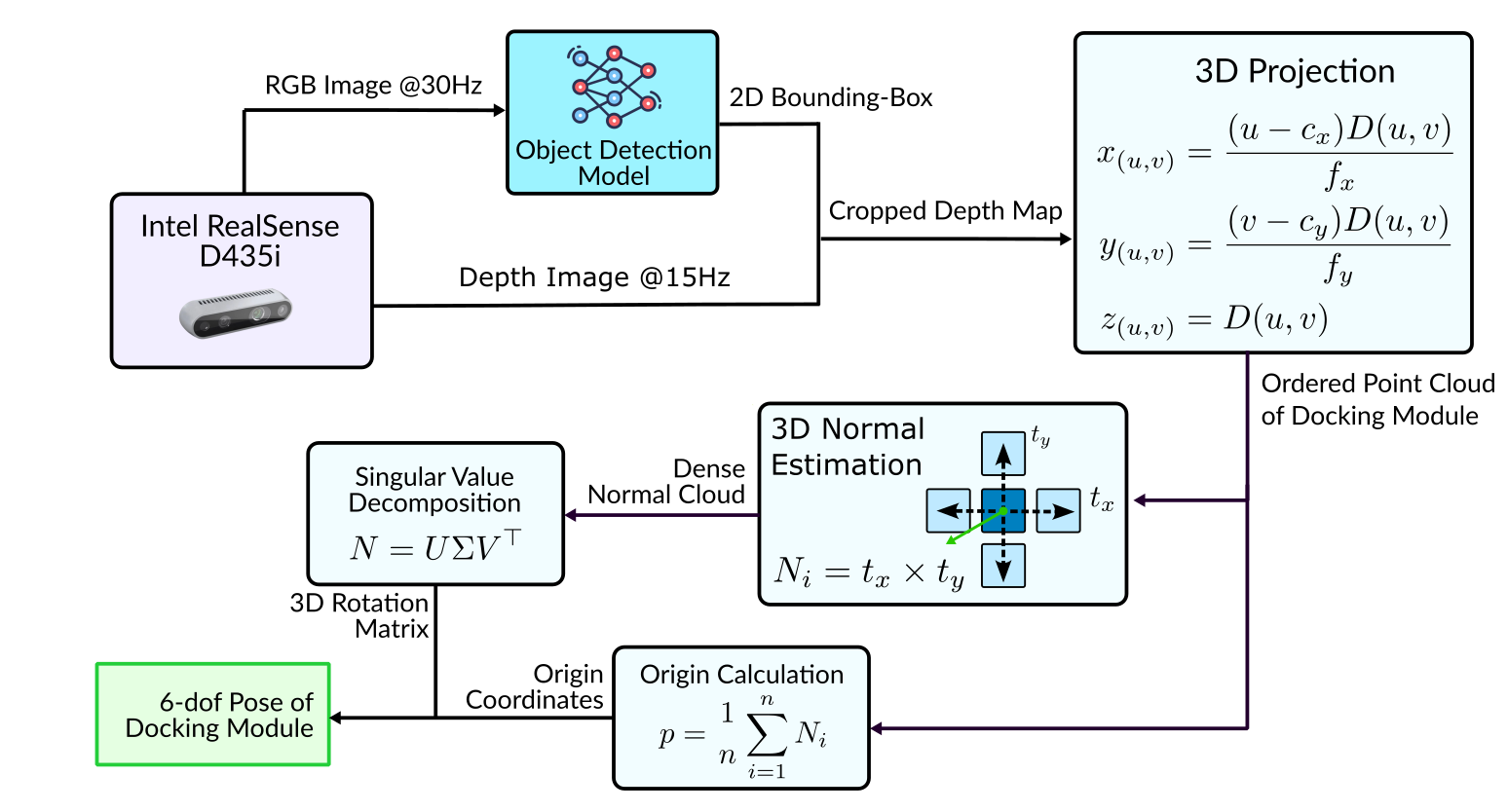}
    \caption{Perception system overview}
    \label{fig:perception-system-overview}
    \vspace{-6mm}
\end{figure}

\section{Docking Module Detection}
\label{sec:detection}

The autonomy stack considered in this study is required to identify a box positioned 5-6 meters away from COBRA. Following the detection, path planning and trajectory tracking are initiated to bring the system into proximity with the box, after which the manipulation phase begins. At this stage, COBRA estimates the 6-DoF posture of both the box and the docking module.

To achieve object detection and segmentation, we evaluated two deep learning-based models: YOLO (You Only Look Once) \cite{jocher_ultralytics_2023} and Mask2Former \cite{cheng_masked-attention_2022}. While both are widely used in computer vision, they address different tasks with distinct methodologies.

OLO is primarily used for object detection, framing it as a single regression problem by predicting bounding box coordinates and class probabilities directly from image pixels. Unlike traditional methods like R-CNN, which process multiple image scales, YOLO processes the entire image in a single pass, offering exceptional speed. In this study, we implement YOLO-v8, which is well-suited for real-time detection given COBRA's computational limits.

While YOLO’s speed is advantageous, it has limitations. For distant objects, like our target box, its grid-based prediction approach can lead to coarse localization. Additionally, in complex backgrounds, YOLO struggles with accuracy, especially for small or overlapping objects.

We also evaluated Mask2Former, a transformer-based model designed for segmentation tasks. By using attention mechanisms, it captures global dependencies within an image, enabling more precise object boundaries than YOLO's bounding boxes. Mask2Former is highly effective in densely packed scenes or when dealing with overlapping objects, making it better suited for complex environments.

Mask2Former generates masks for objects or segments in an image, providing more precise boundaries than YOLO’s bounding box approach. Its transformer-based architecture allows it to perform multiple segmentation tasks with high accuracy, making it more effective in scenes with overlapping or densely packed objects.

\begin{figure}
    \centering
    \includegraphics[width=0.6\linewidth]{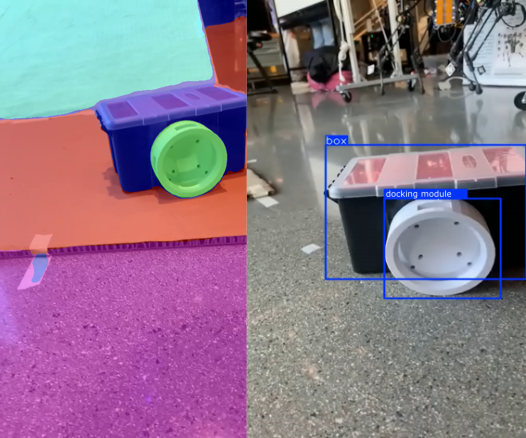}
    \caption{Shows detection of docking module using Mask2Former (left) and Yolo-v8 (right)}
    \label{fig:yolo-mask2former}
    \vspace{-3mm}
\end{figure}

We found that Mask2Former outperformed YOLO in terms of flexibility, accuracy, and handling complex scenes. Its ability to produce fine-grained masks for objects and regions, particularly in cluttered environments, is a key strength. The attention mechanism enables Mask2Former to differentiate between overlapping objects better than traditional convolutional neural networks (CNNs). Figure \ref{fig:yolo-mask2former} shows a comparison of the detection output from Mask2Former and Yolo-v8, highlighting this difference in precision. 

However, speed emerged as a significant limitation of Mask2Former. Segmentation tasks are inherently more computationally demanding than object detection, and Mask2Former was noticeably slower than YOLO during our evaluations. This posed a challenge for COBRA, as its gait cycles last 1-2 seconds, during which an updated estimate is required to ensure the object remains in view. The limited space available for more powerful computing hardware further complicates the integration of more resource-intensive models like Mask2Former into the autonomy stack.

\subsection{Yolo-v8 Training Process}
\begin{figure}
    \centering
    \includegraphics[width=0.8\linewidth]{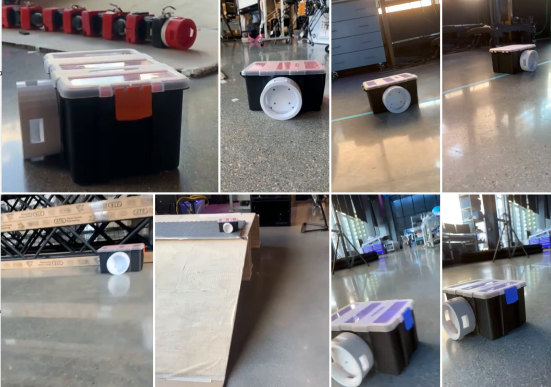}
    \caption{Examples of images from dataset at various ranges from camera}
    \label{fig:dataset}
    \vspace{-5mm}
\end{figure}
We implemented our system using the pre-trained YOLO-v8 Nano model, selected for its lightweight architecture and suitability for real-time detection given COBRA's computational limits. The model was fine-tuned on a custom dataset we curated, featuring various images of the docking module captured under different backgrounds, camera orientations, and distances. Figure \ref{fig:dataset}  shows a subset of this dataset, which includes 570 manually annotated images to accurately label the box and docking module, ensuring precise object localization essential for COBRA's docking accuracy.

After collecting and annotating the dataset, we fine-tuned the YOLO-v8 Nano model, optimizing it to detect both the box and docking module under relevant conditions. Table \ref{tab:yolo-perf}  details the model's performance metrics, including precision, recall, and mean Average Precision (mAP) scores for both object classes. Precision measures the ratio of correctly classified instances (true positives) to all predicted instances, while recall quantifies the proportion of true positives out of all actual instances, indicating whether the model missed any detections. mAP aggregates precision and recall across different detection thresholds, offering a comprehensive metric that reflects both accuracy and the model's ability to avoid missed detections.

\begin{table}[]
    \centering
    \begin{tabular}{c|c|c|c}
        \textbf{Class} & \textbf{Precision} & \textbf{Recall} & \textbf{mAP} \\
        \hline \\
       Box  &  0.88 & 0.84 & 0.68\\
       Docking Module  & 0.93 & 0.67 & 0.50
    \end{tabular}
    \caption{Performance of trained Yolo-v8 Nano model}
    \label{tab:yolo-perf}
    \vspace{-10mm}
\end{table}

For the box class, the model achieved a precision of 0.88, a recall of 0.84, and a mAP of 0.68, indicating strong performance with few false positives. In contrast, the docking module class had a higher precision of 0.93, correctly classifying most instances, but a lower recall of 0.67, suggesting missed detections. This lower recall likely stems from the docking module's varying appearance across orientations and distances, leading to a mAP of 0.50.

While these results are promising, the lower recall for the docking module suggests that further data augmentation or model optimization could improve detection in challenging scenarios like extreme angles or occlusions. The mAP disparity between the box and docking module classes also highlights the model's sensitivity to object characteristics like size and texture.



\section{Pose Estimation of Docking Module}
\label{sec:dock-pose-estim}

In this section, we elucidate the approach we adopted to estimate the 6-DoF pose of the docking module before manipulating the object.

The perception framework presented in Fig. \ref{fig:perception-system-overview} leverages RGB and depth images from an Intel RealSense D435i camera to estimate the 6-DoF pose of a docking module. This system integrates deep learning and geometric estimation to identify both the center and normal vector of the module.

The Intel RealSense D435i provides synchronized RGB (30Hz) and depth images (15Hz), which serve as input. An object detection model based on YOLOv8 processes the RGB image to generate a 2D bounding box around the docking module. The corresponding depth data, extracted using this bounding box, is then used to project pixel coordinates from 2D image space into 3D world space via the camera's intrinsic parameters. The 3D point coordinates for each pixel $(u,v)$ are derived from pixel depth values $D(u,v)$ using 3D projection:
\begin{equation}
\mathbf{p}_{(u,v)} = \begin{bmatrix} 
x_{(u,v)} \\ 
y_{(u,v)} \\ 
z_{(u,v)} 
\end{bmatrix}
=
\begin{bmatrix} 
\frac{(u - c_x) D(u,v)}{f_x} \\ 
\frac{(v - c_y) D(u,v)}{f_y} \\ 
D(u,v) 
\end{bmatrix}
\end{equation}
where $[c_x,~c_y,~f_x,~f_y]$ are camera intrinsic parameters. This results in an ordered 3D point cloud corresponding to the docking module's surface.

To estimate surface orientation, 3D normal vectors are computed from the point cloud. Local tangent vectors $\mathbf{t}_x$ and $\mathbf{t}_y$ are computed at each point by taking the vector to neighboring points, and their cross product produces the normal vector $N_i$ at each point:
\begin{equation}
\mathbf{N}_i = \mathbf{t}_x \times \mathbf{t}_y    
\end{equation}

The normals form a dense normal cloud, which is then processed using Singular Value Decomposition (SVD) to calculate the 3D rotation matrix $\mathbf{R}$ of the module. 
\begin{equation}
\mathbf{N} = \mathbf{U}\mathbf{\Sigma} \mathbf{V}^\top
\end{equation}
here, $\mathbf{U} = \begin{bmatrix}\hat x & \hat y & \hat z\end{bmatrix}$ provides the orthogonal basis vectors of the row space of $\mathbf{N}$. However, without any constraints, the axes derived from $\mathbf{U}$ may not adhere to the right-hand rule, nor do they guarantee stable orientation. To ensure consistency with previous time steps, the directions of $\hat x$ and $\hat z$ are flipped if necessary to align with the previous orientation, and the cross product is used to compute $\hat y$. This process produces the rotation matrix $\mathbf{R}$ that defines the orientation of the docking module with respect to the camera.

To calculate the center of the docking module, we compute the average of the points in the point cloud:
\begin{equation}
\mathbf{p} = \frac{1}{n}\sum_{i=1}^n [x_i,~y_i,~z_i]^\top
\end{equation}

This provides the full 6-DoF pose of the docking module, capturing both its position and orientation in the camera frame in real-time. We then use the pose of the camera obtained from visual inertial odometry to transform the pose into the world frame for loco-manipulation tasks.

\section{Closed-Loop Path Tracking and Docking }
\label{sec:control}

In this section, we briefly explain the path tracking and docking controllers. Snake robot gaits typically involve oscillatory motions across all links. To describe the motion of the entire robot in a way that minimizes the influence of these oscillations, a floating reference frame is required—one that remains mostly stable and independent of the individual link movements. Following the approach in \cite{rollinson_virtual_2012}, we define a center of mass (COM) frame as this floating reference frame. The position of the COM in world coordinates is calculated as:
\begin{equation}
    \mathbf{P}_\text{com} = \frac{\sum_{i=1}^{N} m_i \vec{p}_i}{\sum_{i=1}^{N} m_i}
\end{equation}
where $m_i$ and $p_i$ denote the mass and position of the $i^\text{th}$ link. We stack all $\vec{p}_i$ link positions and form a wide matrix called $\mathbf{P}_\text{links}$. Then, we form another wide matrix $\tilde{\mathbf{P}}_\text{links}$ where the entries are the relative position of the links with respect to the snake COM $\vec p_\text{com}$ given by
\begin{equation}
    \tilde{\mathbf{P}}_\text{links} = \mathbf{P}_\text{links} -\left[\mathbf{P}^\top_\text{com},\dots,\mathbf{P}^\top_\text{com}\right]
\end{equation}
The orientation of the center of mass frame is determined using Singular Value Decomposition (SVD):
\begin{equation}
    \tilde{\mathbf{P}}_\text{links} = \mathbf{U}\mathbf{\Sigma} \mathbf{V}^\top
\end{equation}
Here, $\mathbf{U} = \begin{bmatrix}\hat x & \hat y & \hat z\end{bmatrix}$ provides the orthogonal basis vectors of the row space of $\tilde{\mathbf{P}}_\text{links}$. 

Now, we steer $\textbf{U}$ to desired waypoints $\textbf{U}_{d}$ in the context of nonlinear programming problem (NLP) subject to mainly flat ground kinematic constraints, namely
\begin{equation}
\begin{array}{cc}
\begin{array}{c}
\text{minimize:} \\
\mathbf{X}
\end{array}
&
\mathcal{J}(\mathbf{X)} = \sum_{i=1}^N \lvert\lvert \mathbf{P}_\text{com}(t_i) - \mathbf{P}_\text{com}^\text{des} \rvert\rvert^2
\\
\begin{array}{c}
\text{subject to:}
\end{array}
&
\begin{array}{rcl}
  \mathbf{X}_\text{min} \le &  \mathbf{X}   & \le \mathbf{X}_\text{max} \\
  0 \le &P_\text{com}^\text{pred} - P_\text{com}& \le 0
\end{array}
\end{array}
\end{equation}
where $\mathbf{X} = \begin{bmatrix} 
a & \omega & \varphi & \mathbf{P}_{\text{com}, 1} & \mathbf{P}_{\text{com}, 2} & \dots & \mathbf{P}_{\text{com}, n}
\end{bmatrix}^\top$. 

\section{Results}
\label{sec:res}
\begin{figure}
    \centering
    \includegraphics[width=0.8\linewidth]{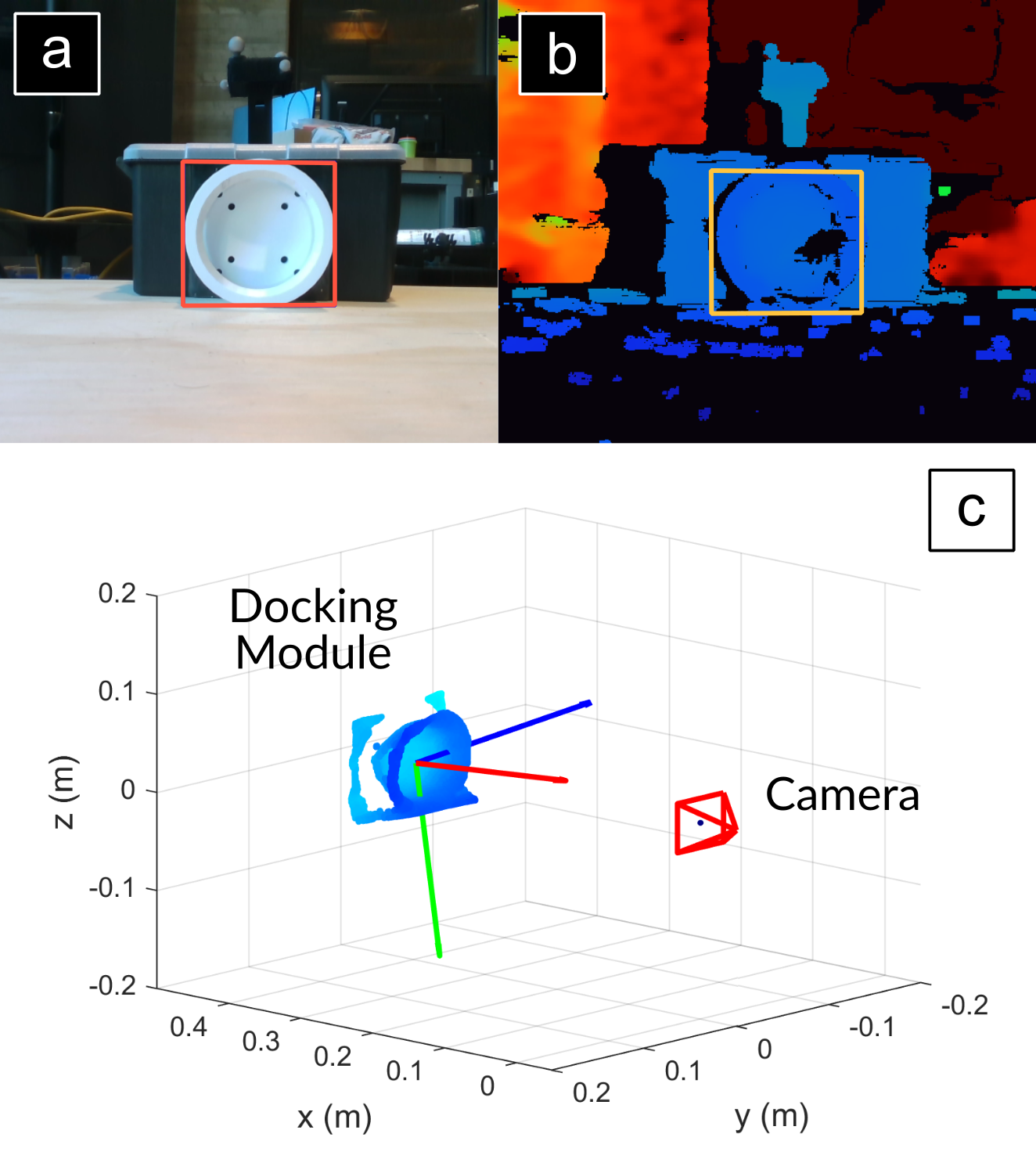}
    \caption{(a) Detection of the docking module in RGB image. (b) Mapping detected region to corresponding depth image. (c) Projected point cloud and SVD based estimation of center and axes of docking module}
    \label{fig:estimation-result}
\end{figure}

We implemented the pipeline presented in Section \ref{sec:dock-pose-estim} onboard COBRA's perception unit, integrating real-time object detection and pose estimation. Figure \ref{fig:estimation-result}  illustrates the outputs generated by this pipeline. In the top-left view, the docking module is detected using the RGB image, with a bounding box accurately framing the module. The corresponding bounding box mapped onto the depth image is shown in the top-right view, providing additional spatial information about the module’s position relative to the camera. Below, we see the projected point cloud of the docking module as well as the center and axes in the camera frame using SVD. 

\begin{figure}
    \centering
    \includegraphics[width=0.9\linewidth]{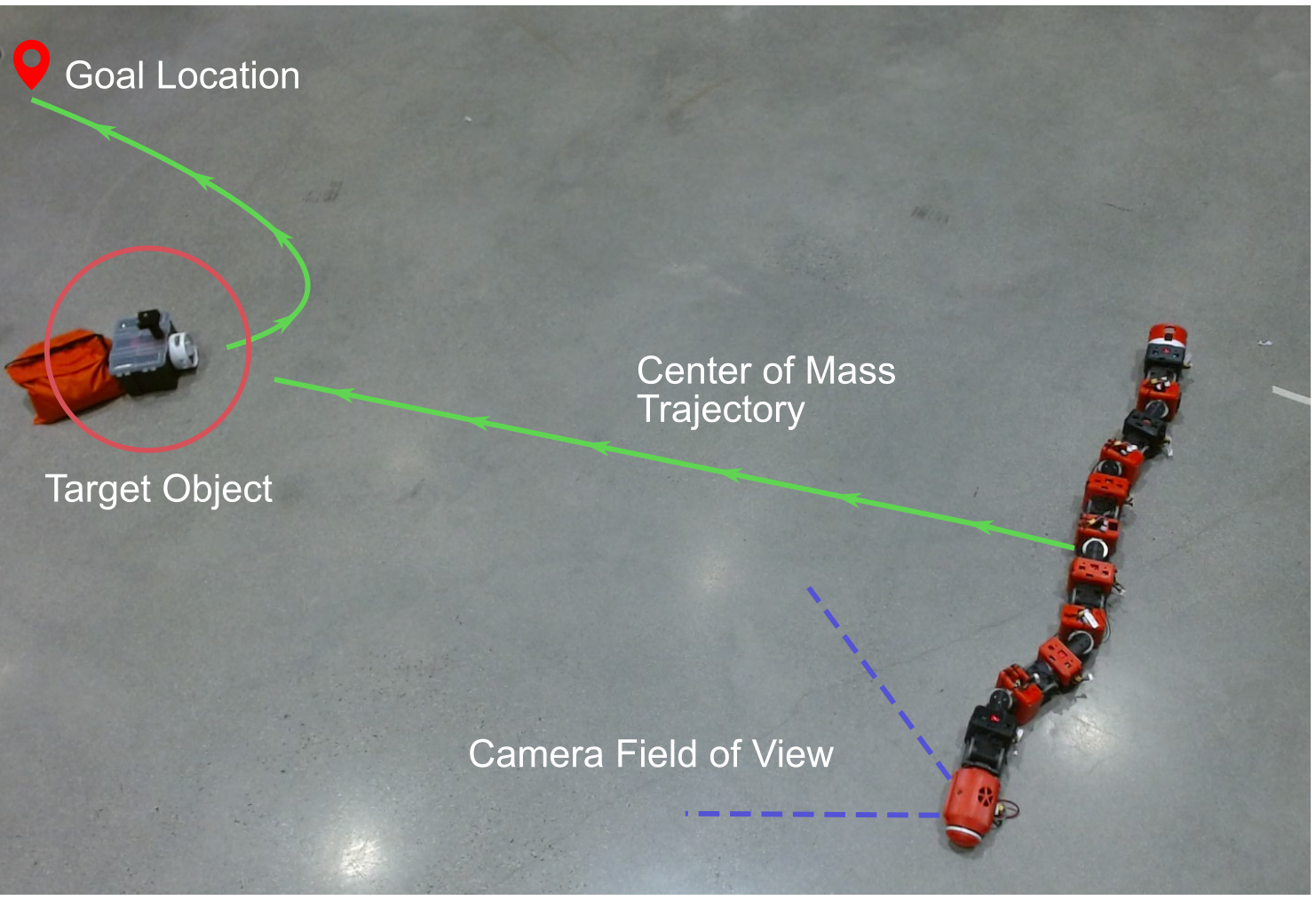}
    \caption{Experiment setup shows the starting position of robot and target box, and the goal position to move the box to}
    \label{fig:exp-setup}
    \vspace{-5mm}
\end{figure}

\begin{figure*}
    \centering    \includegraphics[width=0.9\linewidth]{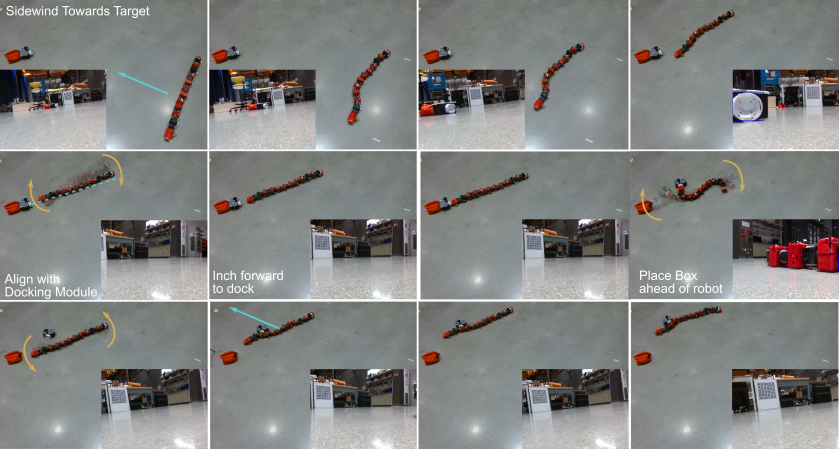}
    \caption{Experiments with COBRA showing loco-manipulation while running detection algorithm onboard (inset: camera view with detection running)}
    \label{fig:snapshots}
    \vspace{-5mm}
\end{figure*}

To validate the functionality of this pipeline, we implement a full loco-manipulation application open-loop according to the experiment setup shown in Figure \ref{fig:exp-setup}. COBRA starts at a distance of 4-5m from the box with the box in view, and sidewinds towards it while continuously running the object detection algorithm onboard at 10 Hz. Once it reaches the box, it uses a predefined set of gaits to dock to the box, place the box ahead of the robot and undock. Finally, COBRA sidewinds to move the box to the final goal location. Figure \ref{fig:snapshots} presents a series of snapshots depicting various stages of the experiment as performed on the physical robot. Inset is the view from the onboard camera on COBRA with the detection running.

\section{Concluding Remarks}
This work demonstrates the successful integration of a YOLO-v8-based perception pipeline and a loco-manipulation framework on COBRA. Through real-time object detection, 6-DoF pose estimation, and open-loop manipulation, COBRA executed a full loco-manipulation task. While the system showed promising results, challenges remain in improving detection accuracy and manipulation robustness. Future efforts will focus on optimizing the perception pipeline, incorporating feedback control, and enhancing COBRA's adaptability to dynamic and cluttered environments.
\balance{}
\printbibliography

\end{document}